\documentclass[runningheads]{classfile}

\usepackage[T1]{fontenc}

\usepackage{graphicx,booktabs}
\usepackage{hyperref}

\usepackage{amsmath,amsfonts,bm}
\usepackage{bbm}
\usepackage{algorithm}
\usepackage{algpseudocode}

\def\1{\bm{1}}

\def\vone{{\bm{1}}}
\def\vI{{\bm{I}}}

\def\vx{{\bm{x}}}
\def\vy{{\bm{y}}}

\DeclareMathAlphabet{\mathsfit}{\encodingdefault}{\sfdefault}{m}{sl}
\SetMathAlphabet{\mathsfit}{bold}{\encodingdefault}{\sfdefault}{bx}{n}

\def\gO{{\mathcal{O}}}

\newcommand{\R}{\mathbb{R}}

\newcommand{\softmax}{\mathrm{softmax}}

\newcommand{\icol}[1]{
  \left(\begin{smallmatrix}#1\end{smallmatrix}\right)%
}

\algblock{Input}{EndInput}
\algnotext{EndInput}
\algblock{Output}{EndOutput}
\algnotext{EndOutput}

\begin{document}

\title{On the Approximation of Phylogenetic Distance Functions by Artificial Neural Networks}

\titlerunning{Approximating Distance Functions with Neural Networks}

\author{Benjamin K. Rosenzweig* \and Matthew W. Hahn}
\authorrunning{Rosenzweig and Hahn}
\institute{Indiana University, Bloomington, IN 47405, USA}

\maketitle             

\begin{abstract}
Inferring the phylogenetic relationships among a sample of organisms is a fundamental problem in modern biology. While distance-based hierarchical clustering algorithms achieved early success on this task, these have been supplanted by Bayesian and maximum likelihood search procedures based on complex models of molecular evolution. In this work we describe minimal neural network architectures that can approximate classic phylogenetic distance functions and the properties required to learn distances under a variety of molecular evolutionary models. In contrast to model-based inference (and recently proposed model-free convolutional and transformer networks), these architectures have a small computational footprint and are scalable to large numbers of taxa and molecular characters. The learned distance functions generalize well and, given an appropriate training dataset, achieve results comparable to state-of-the art inference methods.

\end{abstract}

\section{Introduction}
\subsection{Supervised Machine Learning in Phylogenetics}

In many applications of machine learning (ML), success depends on collecting high volumes of real-world training data and deploying high-capacity networks to uncover complex hidden structures from that data \cite{Vaswani2017a}.  Phylogenetics differs from such domains in the lack of ground truth data.  Instead, researchers assume a generative model that produces the observed diversity among species in the present day.  Any supervised ML method for phylogenetic inference must therefore utilize simulated data for both training and evaluation.  The choice of simulation regime has profound consequences for the performance of a model and its generalizability.

Supervised ML algorithms that can infer trees have thus proved elusive.  \cite{Suvorov2020}, \cite{Zou2019}, and \cite{wang_fusang_2023} used simulated sequence data to train CNN, residual networks, and recurrent neural networks, respectively, to infer an unweighted topology for a single non-recombining locus with $n=4$ taxa. 
As there are only 3 possible unrooted topologies on $4$ taxa, these authors were able to build networks with a multiclass output.  Since the number of unrooted binary trees on $n$ leaves grows as $(2n-5)!!$, this approach is impractical for $n>4$. Quartet-based methods in theory can be combined with a quartet amalgamation algorithm to recover a tree by inferring trees for all quartet subsets of taxa, then building a Maximum Quartet Consistency (MQC) tree (e.g.  \cite{snir2012}).  However, MQC is also an NP-hard problem \cite{steel_complexity_1992} (and is not guaranteed to match the true tree) so little is gained by this approach.

The ordering of taxa in an alignment should not affect the predicted tree; however, CNNs and ResNets are not invariant to permutations of their input.  \cite{Zou2019} use data augmentation to mitigate this problem, a strategy that is not practical for large $n$.  \cite{Solis-Lemus2022} introduced a parameter-tying scheme which ensures that networks are invariant to taxa ordering. However, this scheme requires a unique architecture for each $n$ and is thus also impractical for large $n$.

Because any supervised ML approach to phylogenetic inference must utilize simulated data, the complexity of the assumed generative model should determine the model architecture.  Prior work has not always followed this recipe. \cite{Suvorov2020} trained a CNN on alignments of exchangeable (but spatially uncorrelated) characters.  By contrast, the generative model in \cite{nesterenko2025} included insertions and deletions but their self-attention architecture lacked positional embedding.  In the first case, the architecture had sufficient capacity to capture the spatial features of real-world data, but generalization was limited by the simplicity of the training distribution.  In the second case, the training distribution had spatial dependencies but the architecture had no way to capture these features.

Many groups report results on simulated data that are comparable to maximum likelihood methods.  However, test conditions typically closely resemble the training conditions, and generally comprise fixed-length, non-recombinant, perfectly aligned sequences.  
 In a large comparative study, \cite{Zaharias2022} found that, far from beating maximum likelihood inference, no pretrained DNN was able to consistently match  neighbor-joining with Hamming distances.

\subsection{Metric Learning}
The models we develop here are distance-based: each alignment is transformed into a set of pairwise of distances which are converted into a tree by neighbor-joining (NJ).  One motivation for this approach is computational: this is the only way inference can be done on large ($n>15, L>200$ bp) phylogenies. A second motivation is that with a distance-based model we can begin to leverage theoretical results on the biological assumptions underpinning distance methods.  This approach, we hope, will pave the way for general-purpose, transferrable ML techniques in phylogenetics.

Our work is similar to deep metric learning applications from the computer vision literature \cite{Hu2016a,Yi2014,Wen2019}.  In these tasks, a DNN is trained to predict the 3-dimensional geometry of a scene from a 2-dimensional image.
In those tasks the learned metric space corresponds to physical reality.   In our task the learned metric space should contain the true species tree.  On any given input sample, a good DNN must select between different evolutionary models induced during training, and output an approximate distance matrix.  The quality of this approximation depends on the properties of the NJ algorithm.

Compared to probabilistic phylogenetic methods, metric learning requires minimal computational overhead, as the computation has been offloaded into the model training phase. Metric learning is also useful in situations where the appropriate evolutionary model is difficult to define. 

The most general architectures we explore rely on attention \cite{Vaswani2017a}. 
Recently, \cite{nesterenko2025} proposed Phyloformer, a transformer architecture similar to our own, based on the same principle of information sharing across taxa and sites.  Where appropriate, we include their pretrained models among our evaluations.  

\subsection{Symmetry-Preserving Deep Learning}

Convolutional neural networks have achieved considerable success in both computer vision and natural language processing \cite{Szegedy2015}.  Over the last decade researchers in the field of `geometric deep learning' have expanded this concept to a larger family of neural network layers that are invariant to other group actions beside translations.  These include rotations, permutations \cite{Zaheer2017,lee_set_2018}, and translations on compact manifolds \cite{bloem-reddy2020}.  The motivations for all such approaches are the same: first, by ignoring symmetries in the input data, the number of parameters required by the model can be dramatically reduced. Second, the model is guaranteed to behave equivalently on equivalent inputs, and this stability is achieved without resorting to costly data augmentation.

Invariance is an essential property for any end-to-end phylogenetic inference algorithm; every permutation of the $n$ input sequences must result in an equivalent permutation of the $n$ labels in the output tree.  Without this, a network would need to learn these $n!$ symmetries from data.  

For $n>4$, a network must be of very high capacity in order to ``memorize'' equivalent inputs, and the training regime would require a huge amount of data augmentation.  Such approaches are practical in computer vision \cite{silver_mastering_2017,shorten2019}, where the number of rotational and reflectional symmetries is small and independent of the input size.  This is not the case for a structure learning problem where the number of symmetries grows with $n!$.
We therefore explore a class of permutation equivariant neural network layers that are appropriate for multi-sequence alignments.

\subsubsection{Spatially Correlated Characters}

While exchangeability of sequences is crucial for a phylogenetic DNN, exchangeability of sites may or may not be justified by the evolutionary model.  Many models of molecular evolution assume that sites differ in their rate of evolution.  These sites, while not i.i.d., are still exchangeable \cite{diaconis1980}. A DNN architecture for use with such data should therefore be permutation invariant along the sites dimension.

Other evolutionary models allow sites to be spatially correlated.  These include codon models, models of multinucleotide mutations, and models that allow insertions and deletions.  Inference under such models typically involves pseudo-likelihood optimization \cite{Christensen2005,matsen2022}.
Spatially correlated features can be extracted by a DNN with 1-dimensional convolutions along the sites dimension, or with positional embedding.
More complex correlations between sites arising from conserved motifs can be captured with DNA or protein language models \cite{zhou2024a}.  We evaluate both CNN and positional embedding layers, but the lack of generative \textit{evolutionary} frameworks for DNA and protein language models puts the latter, for the moment, beyond the scope of supervised ML.

\section{Methods}
The simplest distance function on two sequences is the Hamming distance, $d_H$. 
$d_H$ can be applied to both nucleotide and amino acid alignments, and can capture phylogenetic signal at relatively short evolutionary distances.
Methods that account for multiple mutations at the same site over longer time scales are strictly more accurate than $d_H$ at estimating these distances.  The simplest of these, the Jukes-Cantor corrected distance $d_{JC}$, assumes all states are equally frequent and all transitions are equally probable. 

More complex evolutionary models divide nucleotides into rate categories.  Kimura's 2-parameter distance $d_{K2P}$ \cite{Kimura1981}, for example, weights transitions  differently from transversions.
  The JC and K2P models implicitly assume a uniform stationary distribution over bases (i.e. $\pi_A=\pi_T=\pi_C=\pi_G=0.25$).
  K2P also assumes  that transitions are twice as likely to occur as transversions.  The HKY model relaxes these assumptions; the state frequencies and the transition/transversion ratio are unknown parameters of the evolutionary process.
Coestimating pairwise distances allows information about these global parameters to be incorporated into distance estimates, resulting in more accurate trees \cite{nei_molecular_2000,Tamura2004}.  
A neural network approximation of $d_{HKY}$ must be able to compute the frequencies of all four nucleotides in a pair sequence and, for greater accuracy, \textit{across many pairs.}
 
If mutation rates vary between sites, the columns of an alignment are no longer i.i.d; they are merely exchangeable.   
To assign sites to mutation rate categories, a neural network should be able to accumulate information across many sequences in the alignment, similar to the expectation-maximization approach utilized in other distance-based methods.  

Finally, evolutionary models for insertions and deletions (``indels") typically draw gap lengths from a hypergeometric distribution.  This renders the gap characters (and hence the entire sequence) spatially correlated.

In our experiments we evaluate all architectures on JC, K2P, HKY (with Gamma-distributed rate categories) DNA sequences, as well as LG protein sequences with Gamma-distributed rate categories and indels.

\subsection{Network Architectures}
We introduce the following families of architectures:
\begin{enumerate}
    \item \textbf{Sequence networks ($S$)}.  In this architecture, each sequence $\vx_i$ in an alignment $X\subset S^{h\times L}$ is mapped independently to an internal representation $g(\vx_i)\subset \R^{n\times k}$.  The matrix $G=g(\vx_i)_i \subset \R^{n\times k\times L}$ is then modified via attention, convolution, and pooling layers to produce a representation matrix $Z\subset\R^{n \times d}$.  The final output is either:
    \begin{enumerate}
        \item the pairwise Euclidean distances $||Z_i-Z_j||_2$.  We call networks of this class $S_d$.  
        \item  the inner products $Z_i^TZ_j$.  These can be converted to distances for use in NJ by means of the inverse Gromov transform:  $d_{ij}=Z_i^TZ_i+Z_j^T Z_j-2Z_i^TZ_j$.  We call networks of this class $S_c$.  
    \end{enumerate}
    \item  \textbf{Pair networks ($P)$}.  In this architecture, each sequence $\vx_i$ is mapped to a representation $f(\vx_i)$. Each pair of transformed sequences is concatenated to produce a matrix $G'\subset \R^{ {n \choose 2} \times k\times L}$.  $G'$ is then modified via attention, convolution, and pooling to produce a representation matrix $Z\subset\R^{ {n \choose 2} \times k}$.  The final output is then obtained by applying a nonlinear map $g:\R^k\to \R^+$ to each $Z_i$ obtain the ${n \choose 2}$ pairwise distance estimates. 
\end{enumerate}

$S$ networks always produce a valid distance (or covariance), whereas a $P$ network is only guaranteed to produce a dissimilarity.  Furthermore, the space required by an $S$ network is linear in $n$ while that of a $P$ network is quadratic in $n$. $P$ networks, however, are more expressive and easier to train.  
Care must be taken in the construction of an $S$ network to achieve results comparable to $P$ networks; we review some theoretical results that justify this approach below.

 \subsubsection{The Expressive Power of Euclidean Embedding Networks $S_d$}

Here we define the \textit{distortion} of a mapping $f:(X,d_1)\to (Y,d_2)$.  We say that $f$ has distortion $\rho$ if there exists a constant $r>0$ such that for all $x,y\in X$:
\[r d_1(x,y)\leq d_2(f(x),f(y)) \leq r \rho  d_1(x,y)
\]

Bourgain's theorem guarantees that any metric $d$ on a set $X$ of $n$ elements, can be embedded into a $\ell_2$ metric space of dimension $k=\gO(\log^2(n))$ with distortion $\rho=\gO(\log(n))$ \cite{bourgain_lipschitz_1985}.  The Johnson-Lindenstrauss Lemma guarantees that this can be reduced to dimension $k=\gO(\log(n))$.  \cite{linial_geometry_1995} give an explicit randomized algorithm for $\ell_2$ embeddings that achieve the minimal distortion for any metric $d_1$.  This algorithm selects a sequence of random subsets $A_i\subset X, |A_i|=2^i,1\leq i\leq k = \lfloor \log(n)\rfloor$.  Then each point $x$ is sent to the vector $(d_1(x,A_1),...,d_1(x,A_k))$, where $d_1(x,A)\equiv \min_{y\in A}d_1(x,y)$. As noted in \cite{linial_geometry_1995}, truncating the vectors at some $i<k$ has the effect of pulling nearby points in $d_1$ closer together, with no effect on distant points.  By choosing a Euclidean space of sufficient dimensionality, we can minimize the distortion and hope to obtain a reasonable approximation of the true tree metric.

This suggests several considerations for designing a DNN to approximate $d_T$ via sequence embeddings in $\ell_2^k$: (1) the embedding dimension $k$ is a crucial factor, and will depend on the maximum number of taxa $n$ (not the sequence length $L$) and (2) the optimal embedding must be able to share \textit{some} information between sequences; this is true even for the simplest generative model, JC. 

\subsubsection{The Expressive Power of Inner Product Networks $S_c$}

Every rooted tree $T$ can be identified with both the matrix of pairwise tree distances $D^T$ and the matrix of phylogenetic covariances $C$, where $C_{ab}$ is the shared distance from taxa $a$ and $b$ to the tree root $\rho$.  If covariances are learned with sufficient fidelity, a tree can even be constructed without utilizing NJ.  The root of the tree can also be recovered, something that is impossible for any distance matrix method.
In addition, $C$ can be learned without sharing any information between sequences \textit{provided that the final embedding is of sufficient dimensionality.}  This is a consequence of Mercer's theorem \cite{Kulis2009,kulis2013}.

\subsection{Permutation
Invariant Networks}

A distance function $d:\mathcal{S}^L\times \mathcal{S}^L\to\R$ on sequences $(s_0\cdots s_1)\in \mathcal{S}^L$ is invariant under joint permutations of its arguments (a pair of points in the metric space) and features (the coordinates $i=1,...,L$).  In other words, it is invariant under the action of the direct product group $\mathbf{S}_L\times \mathbf{S}_2$.   

Thus, \textit{if} a pairwise distance function $d(\vx,\vy)$ can be approximated by a neural network (with the approximation being arbitrarily precise as the number of parameters $\to\infty$), the network $f(\vx|\vy)$ has the following form: \begin{align}
    f(\vx|\vy) = g(f_{W_n}(\cdots \sigma( 
 f_{W_1} (\vx|\vy))) 
\label{eq:equivariant}
\end{align}
 Here the output layer $g:\R\to\R$ is an arbitrary neural network $z\mapsto\sigma(V_{m}\sigma(\cdots V_1 z))$ capturing any nonlinear scaling of the distance. $f_{W_n}(\vx|\vy)=w_1 \sum_{i=1}^L (x_i+ y_i)+w_2$ is a final invariant layer, and the layers $f_{W_1},...,f_{W_{n-1}}$ are equivariant layers.  Each equivariant weight matrix $W$ has the form\footnote{for convenience we add a bias term $w_5$.}

\begin{align}
    W=w_1\vI_{2d}+ w_2 \begin{bmatrix}
\mathbf{0} & \vI_d\\
\vI_d & \mathbf{0}
\end{bmatrix}	  + w_3    \begin{bmatrix}
\vone_{d}\vone_{d}^T & \mathbf{0} \\
 \mathbf{0} & \vone_{d}\vone_{d}^T
\end{bmatrix}	+ w_4
    \begin{bmatrix}
\mathbf{0} & \vone_{d}\vone_{d}^T\\
\vone_{d}\vone_{d}^T & \mathbf{0}
\end{bmatrix} +w_5
\end{align}\label{eq:equiv-layer}

To accumulate more information (such as stationary frequencies), additional ``heads'' (parallel equivariant layers) can be added.  This architecture is sufficient to approximate a distance function on sequences where all sites are i.i.d. 

\subsubsection{Spatially Correlated Characters}

CNNs are invariant to translations of input features, thus making them appropriate for gapped alignments.  The invariant architectures described above can be easily adapted to spatially correlated data by adding a 1-dimensional CNN along the $L$ dimension.
Transformers can also be used on spatial data: self-attention by itself ignores location, but spatial features can be extract by means of positional encoding, or by a variety of local attention mechanisms (see Appendix, Supplementary Methods).

\subsection{Taxa-wise Attention: Learning Graphs on the Fly}

If something about the relationships between taxa were known \textit{a priori}, distances can be estimated with greater accuracy \cite{nei_variances_1989}.  
 Above we showed that an equivariant linear transformation for pairs of sequences can be represented with just four parameters. Using similar techniques, \cite{Maron2019} derive a basis for equivariant linear transformations on  arbitrary graphs which requires just 15 parameters (or 6 if the input graph is symmetric and lacks self-loops). 
 
 Of course, if the relationships between taxa were known, the phylogeny would already be solved.  
Here self-attention offers us the chance to learn relationships between sequences on the fly and to update their embeddings accordingly.  \cite{Kim2022} proved that self-attention can approximate any graph equivariant layer, provided each attention layer has at least 15 heads and is followed by an MLP layer.  In our application, we can apply taxa-wise attention both to the $n\times k\times L$ dimensional alignment (an S network), as well as the expanded ${n \choose 2} \times k \times L$ matrix of sequence pairs (a P network).  While the latter is rich enough to represent features of any graph, the former is more memory efficient and still shares enough information between sequences to reconstruct phylogenies.  In our experiments we compare site-attention and full site+taxa attention on S, P, and hybrid SP networks.

\section{Results}
\label{representing}

\subsection{Approximating $d_H,d_{JC},$ and $d_{K2P}$}
A neural network with at least two hidden layers can always approximate the Hamming distance.  This follows from a famous argument by \cite{Minsky1988}: approximating a Hamming distance amounts to approximating the XOR function.    
To see this, suppose that $x,y\in\{0,1\}^{k\times L}$ are two  matrices representing strings of discrete characters in some space $S$ using a one-hot encoding of dimension $k=|S|$.  Their normalized Hamming distance can be represented by a 4-layer equivariant network in the form of Equation \ref{eq:equivariant} with ReLU activations.  The weight vectors of its layers are given by $W_1=(2,0,0,0,1)^T$, $W_2=(1/2,0,0,0,0)^T$, and $W_3=(1/L,0)^T$ (the invariant layer).  The final layer $g$ is simply a summation over the $k$ channels.

By similar reasoning, $d_{JC}$ can also be represented by modifying the final $g$ network to approximate $-\frac{3}{4}\ln(1 - \frac{4}{3}z)$. $d_{K2P}$ can be represented as follows.  After $f_{W_2}$ there are 2 vectors corresponding to transitions and 4 corresponding to transversions.  For convenience, we alternate equivariant layers with 1-dimensional convolution layers along the $k$ (channel) dimension.     This allows us to add a projection layer $h:\R^4\to\R^2$ after $f_{W_2}$. 
  If $h$ maps transitions to $\icol{1\\0}$ transversions to $\icol{0\\1}$, then applying $f_{W_3}$ gives $\icol{p\\q}$.  Adding a final approximation $g\left( p,q\right) \approx -\frac{1}{2}\ln(1-2p-q) -\frac{1}{4}\ln(1-2q)$ yields the desired result.

Note that for both $d_{JC}$ and $d_{K2P}$ the dimensionality of the network activations decreases from $k\times 2L$ to $k\times 1$ within the first three layers.  Most of the optimization will occur in deeper layers of $g$ as the network learns to approximate simple algebraic functions.

A transformer network can also approximate $d_H$, $d_{JC}$, and $d_{K2P}$. Here alternating self-attention and channel-wise convolution becomes crucial.
Early layers of the network quickly converge to a single representation for each site: $p$ for $d_H$ and $d_{JC}$; $\icol{p\\q}$ for $d_{K2P}$.  In subsequent layers the channel-wise convolutions play the role of the $g$ function, independently transforming each site to a parallel estimate of the target distance.  For this generative model, this representation is clearly wasteful in terms of memory (propagating a redundant activation matrix) and computation (computing $n^2$ attentions between identical sites).

\subsection{Experiments}

We evaluated six basic architectures:
\begin{enumerate}
    \item \textbf{Sites-Invariant-S}: a 2-layer site-invariant sequence network,
    \item \textbf{Taxa-Invariant-S}: a 2-layer site+taxa-invariant sequence network,
    \item \textbf{Sites-Attention-P}: a 6-layer sites-only pair network,
    \item \textbf{Hybrid-Attention-SP}: 3 sites+taxa sequence layers followed by 3 sites-only layers,
    \item \textbf{Full-Attention-S}: a 6-layer sites+taxa sequence network, and
    \item \textbf{Full-Attention-SP}: 3 sites+taxa sequence layers followed by 3 sites+taxa layers.
\end{enumerate} 
  In Table \ref{tab:sizes} we report both model sizes and memory usage for all architectures.  
Self-attention leads to enormous memory requirements for long sequences, even when the number of parameters is small.  
  
\begin{table}
    \centering
    \begin{tabular}{cccc}
     Model    & Trainable Param. & For./back. pass size (MB) & Total Size (MB) \\
     \hline
     Sites-Invariant-S  &7,657  & 103.11 & 103.46 \\

    Full-Invariant-S   & 15,517 & 390.85 & 391.23 \\
    Sites-Attention-P     & 301,285 & 25743.04 &25744.57 \\
    Hybrid-Attention-SP  &355,493  &14560.96  & 14562.70 \\
   Full-Attention-S   & 401,125 &3987.84  & 3989.76\\
   Full-Attention-SP   & 405,413 &20397.76  & 20399.70\\
     Phyloformer    & 308,449 & 28647.36 & 28648.91 \\
    \end{tabular}
    \caption{Model size and memory requirements for single a training step over a batch of 4 alignments of 20 taxa and 1000 sites. Although our two full-attention networks are larger than Phyloformer, they are still more memory efficient for learning and inference.}
    \label{tab:sizes}
\end{table}

By construction, all $P$ networks produced valid dissimilarities.
Surprisingly, output matrices for some trained networks also satisfied the triangle inequality, even when trained without regularization (see Appendix, Table 2).  This behavior appeared within the first 10 epochs and remained stable up to 100 epochs. 

 For i.i.d. data, site attention is simply a form of regularization: site patterns are progressively reduced to a small set of ``representative" patterns. Thus, the final activations of the Sites-Attention-P network have less than $2\%$ of the site patterns as the input DNA sequences (see Appendix, Table 2).  Protein sequences, with $5$ times as many character states as DNA, experience less compression.  The other attention networks incorporate both site and taxa attention and thus experience no compression; each site carries information about sites within a sequence as well as information about the character state at that site in all other sequences.
For Hybrid- and Full-Attention networks, the number of unique site patterns is determined by the first taxa attention layer; this does not change in deeper layers.

Although IQ-TREE outperforms all supervised ML methods in three out of five conditions, the trained networks outperform pairwise distances $d_H,d_{JC},$ and $d_{K2P}$ in the majority of conditions (Table \ref{tab:rf}).
Among the 6 architectures, simpler networks outperformed complex networks on simple evolutionary models; for complex evolutionary models the opposite was true.

The performance of all networks declines sharply when the number of heads are reduced below four.  This is consistent with the theoretical results of \cite{Kim2022} that argue multi-headed attention is crucial for learning graph-structured data.  The number of hidden channels is also crucial: validation error rapidly increases when the number of channels is too low (below 32) or too high (above 128). 
Increasing the size of the training set beyond $10^5$ exempla had no noticeable effect. 
We found that $S_d$ and $S_c$ networks had very similar performance; we report the results of $S_d$ only.  

We found that accuracy increases for all networks as sequence length increases (Appendix, Figure 2). This reassures us that our methods are statistically consistent in the same sense as maximum likelihood distances: they recover the true tree as the number of exchangeable sites goes to infinity.

\begin{table}
  \caption{Mean Robinson-Foulds distance between true and predicted 20-taxa trees.  Complexity increases from left to right and from top to bottom.  For each test condition, the DNN models are trained on in-distribution data.   Likewise IQ-TREE is provided with the correct model specification for likelihood calculations.  The * indicates that spatial layers have been added to the architecture (CNN for Full-Invariant-S, positional encoding for the attention networks).  The best performing model for each condition is bolded. }
  \label{tab:rf}
  \centering
    \begin{tabular}{llllll}
    \toprule
    &&\multicolumn{4}{c}{Mutation Model}                   \\
    Method       &JC      & K2P      & HKY  & LG   & LG+indel\\
    \midrule
    $d_{H}$     & 0.1244  & 0.1204   & 0.1032    & 0.0847 & 0.0739 \\
    $d_{JC}$    & 0.1993  &  0.1475  &0.0889    & 0.0690 & 0.0646 \\
    $d_{K2P}$   & 0.2229  & 0.1963   & 0.0893    & -      & - \\
    Site-Invariant-S & 0.1274  & 0.1389  & 0.1018    & 0.0810  & 0.2087 * \\
    Full-Invariant-S & 0.1253   & 0.1527    &   0.0975  & 0.0695 	& 0.1382 *\\
   
    Sites-Attention-P  &  0.1269  & 0.1219  & 0.1060 & 0.0661 & 0.0716 * \\
    Hybrid-Attention-SP &\textbf{0.1163} & \textbf{0.1011} &  0.0856 & 0.0532     &  0.1299 *  \\
    Full-Attention-S   & 0.1664   & 0.1359 &    0.1222   & 0.0804   & 0.0826 * \\
    Full-Attention-SP   & 0.1664   & 0.1020 &     0.0854  & 0.0547   & 0.0558 * \\
    Phyloformer &  - & -   & -                   & 0.0506 & 0.0601 \\
    IQ-TREE     & 0.1732 & 0.1613   & \textbf{0.0791 }   &  \textbf{0.0438} & \textbf{0.0453} \\
    \bottomrule
  \end{tabular}
\end{table}

\section{Discussion}
Our goal in these experiments was to show the scaling of evolutionary model complexity with neural network complexity.  We therefore trained separate DNN models for each evolutionary model.  For practical applications it will be necessary to train a model with sufficient capacity to learn multiple models and to differentiate between them. 

We appealed to Bourgain's theorem to argue that sharing information among sequences should always be beneficial.  It is only in very degenerate cases---such as a `perfect' phylogeny---that pairwise distances can capture all the information about the internal branches of a tree. As the mutation model increases in complexity, we would expect that the importance of information-sharing across taxa would increase.

It is therefore surprising that $d_{JC}$, Sites-Invariant-S, and Sites-Attention-P methods--which share no information along the taxa dimension-- all perform as well as they do.  In fact, we attempted to train a Taxa-Attention (without site-attention) network and found that it performed abysmally.  That the Hybrid-Attention-SP network frequently exceeds the performance of Full-Attention-SP, however, suggests that only a small amount of information sharing between taxa is actually required.  
This may be related to the greedy nature of NJ: if local relationships among sequences are learned correctly, errors in longer distances are averaged out as each cluster's root is recomputed.

If true, this would suggest that incorporating the tree-building process into the inference procedure could lead to greater efficiency, and that these learned distances---similar to maximum likelihood and maximum parsimony algorithms---are vulnerable to pathological tree shapes (such as long-branch attraction \cite{bergsten2005}).

Real genomic datasets consist of many recombination blocks corresponding to different gene trees.  In preliminary experiments we found that the pretrained networks evaluated above transfer well to concatenated alignments, but fail to match the accuracy of $d_{JC}$ or $d_{K2P}$.   Furthermore, transformer networks cannot \textit{learn} effectively from concatenated alignments.  The large sizes of concatenated alignments (hundreds of times larger than those considered here)  means that some combination of sequence networks, hybrid networks, or pooling will be required.

\section{Data Availability}

All code used in this project is available at \href{https://github.com/bkrosenz/phyloDNN}{github.com/bkrosenz/phyloDNN}.

\begin{credits}

\subsubsection{\discintname}
The authors have no competing interests to declare that are
relevant to the content of this article. 
\end{credits}

\clearpage
\bibliographystyle{bibstyle}
\bibliography{thesis,references}

\appendix
\section{Appendix}

\subsection{Supplementary Tables}

\begin{table}
  \caption{Each cell indicates whether the model is (+) or is not (-) a metric, as well as the compression efficiency of learned site patterns: the number of unique site patterns in the final hidden layer divided by the number of unique site patterns in the input tensor (i.e. $+|1.000$ indicates a learned metric with the same number of site patterns in the input and final representation). }
  \label{tab:metric}
  \centering
    \begin{tabular}{llllll}
    \toprule
    &&\multicolumn{4}{c}{Mutation Model}                   \\
    Method            &JC     & K2P     & HKY     & LG    & LG+indel\\
    Sites-Attention-P    &   $+\,|\, 0.015 $    &  $+\,|\, 0.016 $       & $+\,|\, 0.018 $      & $+\,|\, 0.153 $ & $+\,|\, 0.179 $  \\
    Hybrid-Attention-SP  & $+\,|\,1.000$       & $+\,|\,1.000$   &  $+\,|\,1.000$     & $-\,|\,1.000$    &  $+\,|\,1.000$  \\
    Full-Attention-S  &  $+\,|\,1.000$     &   $+\,|\,1.000$      & $-\,|\, 0.845 $  &$-\,|\,  0.963 $ & $+\,|\,1.000$  \\
    Full-Attention-SP  & $+\,|\,1.000$ &  $+\,|\,1.000$       &     $+\,|\,1.000$       & $-\,|\,1.000$  & $+\,|\,1.000$ \\
    \bottomrule
  \end{tabular}
\end{table}

\clearpage

\subsection{Supplementary Figures}

\begin{figure}    
	\includegraphics[width=\linewidth]{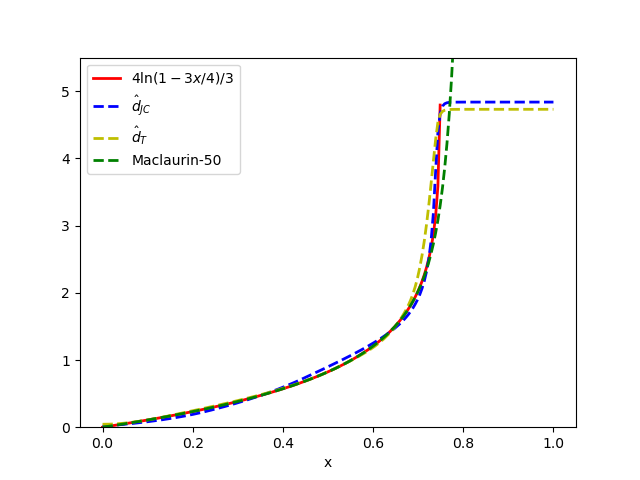}
        \caption{A 6-layer 779-parameter ELU network was trained for 50 epochs on 20-taxa JC alignments with the hamming distance $d_H$ as the function's argument. Both the network $\hat{d}_{JC}$ trained with $d_{JC}$ as the target and the network $\hat{d}_{T}$  trained on the expected divergence $d_T$ closely approximate the function $4\ln(1-3x/4)/3$. Both networks predict a constant value beyond the point $x=0.75$ at which the Jukes-Cantor distance goes to infinity.  This is appropriate behavior for a phylogenetic reconstruction algorithm.  For comparison, the order 50 Maclaurin series (a function with 100 parameters) provides a similar approximation of the log function but rapidly diverges at $x=0.75$.
        The fact that the learned transformations outperform maximum likelihood distances suggests that these transformations convey additional information about the evolutionary process, such as a `prior' on the BD process generating trees in the training set.  In this case, the ceiling learned by $\hat{d}_T$ network is $4.840$, higher than the mean diameter of trees in the training dataset ($3.697$) but considerably less than their maximum diameter ($19.145$).    
    }
    \label{fig:head}
\end{figure}

\begin{figure}
  \centering
      \includegraphics[width=\linewidth]{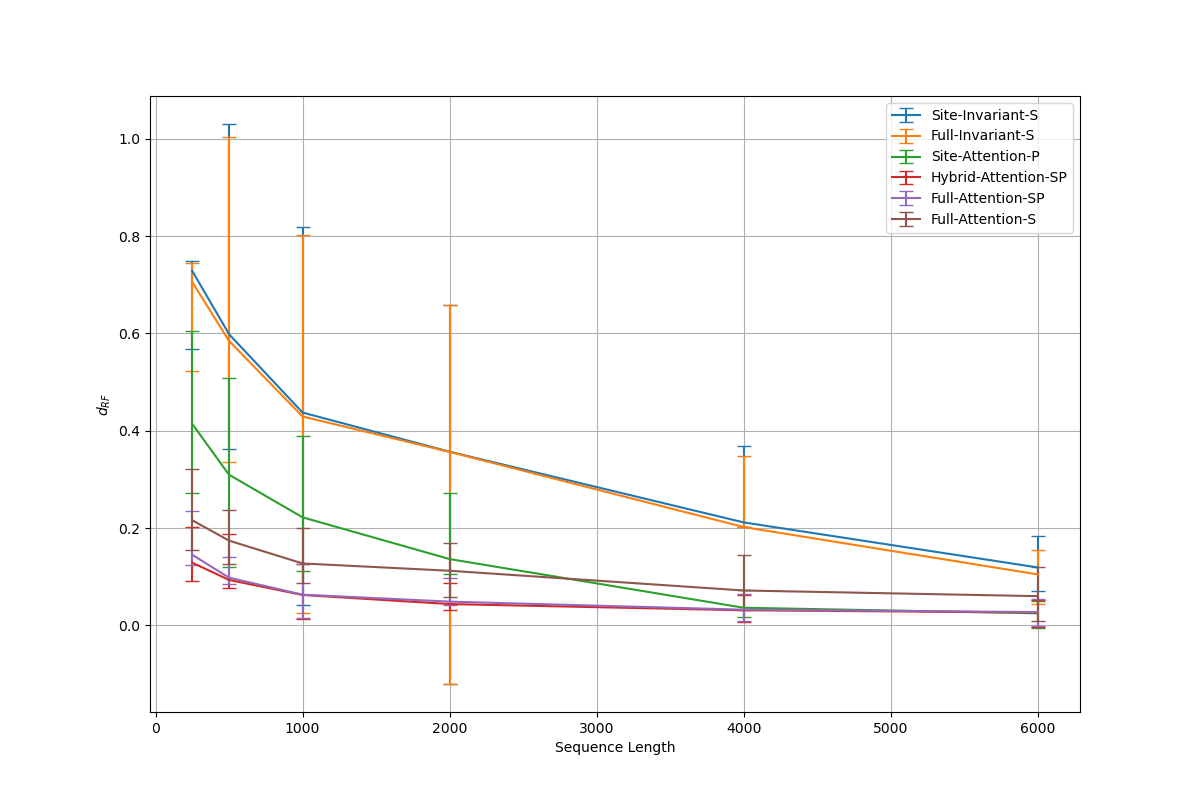}
      \includegraphics[width=\linewidth]{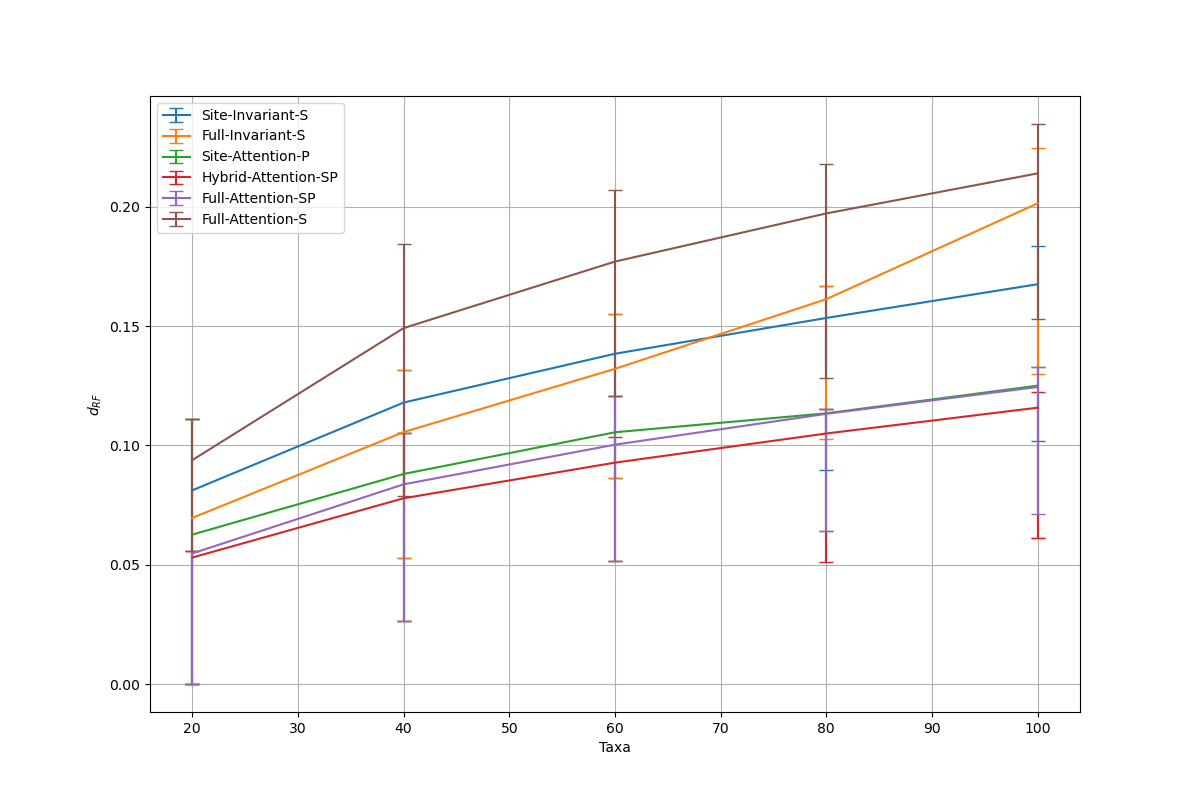}
  \caption{Scaling of performance with input size for LG networks. Top: $d_{RF}$ decreases with sequence length.  Bottom: $d_{RF}$ increases as the number of taxa are increased.  In both figures results are averaged across 500 trees; error bars represent the $50\%$ IQR. }
  \label{fig:lengths}
\end{figure}

\clearpage

\subsection{Supplementary Methods}

\subsection{Self-Attention}

Scaled dot-product attention requires four trainable matrices: $W_q,W_k,W_v\in\R^{d\times \frac{d}{H}}$, where $d$ is the dimension of the latent space and $H$ the number of attention heads.  For a given input $X\in\R^{L\times d}$ the attention block updates the representation of $X$ as follows:
\begin{align*}
    Q_h &\gets XW_q\\
    K_h &\gets XW_k\\
    V_h &\gets XW_v\\
    A_h &\gets \softmax\left(\frac{Q_hK_h^T}{\sqrt{d}}\right)\\
    X &\gets X+\text{Concat}_{h\in[H]} A_hV_h
\end{align*}

The skip connection is necessary to prevent the activations from degenerating to a rank-1 matrix  \cite{Dong2021}.  This block is permutation invariant in the first dimension of $X$: each site is updated based on the value of other sites in the sequence, but these updates depend solely on the  similarity of sites (the argument of the softmax function), not on their indices.

Graph Neural Networks are a static form of attention; in the schema above the $A_h$ matrix is determined by the adjacency matrix of a fixed graph.
Dynamic edge convolutions \cite{wang_dynamic_2019} construct edges from the input data $X$: this is a simple attention mechanism with $W_q=W_k=I$ and a thresholded `hardmax' activation.  

A significant drawback of softmax self-attention is the quadratic dependence on sequence length.  In this work we utilize the linear approximation proposed by \cite{katharopoulos_transformers_2020}.  Various other approaches have been proposed to reduce the time and space complexity of transformers. These include Reformer \cite{kitaev_reformer_2020}, asymmetric clustering \cite{daras_smyrf_2020}, local attention \cite{parmar_image_2018}, compressed attention \cite{liu_non-local_2021}, and attention pooling \cite{zhang_poolingformer_2021}. Commercial large language models increasingly rely on local attention \cite{team_gemma_2025} and Grouped-Query Attention \cite{ainslie_gqa_2023} - parameter sharing across attention heads - to increase the size of input context.  Approaches that reduce the memory footprint will be particularly crucial in extending our results to concatenated alignments with millions of sites.  

\subsection{Simulation Conditions}
Trees were simulated from a birth-death model $BD(\lambda,\mu,n)$ with birth rate $\lambda=1$ death rate $\mu=0.5$ and $n=20$ for training data.  Alignments were simulated using \verb|IQ-TREE|'s AliSim \cite{ly-trong2023}.  The models employed were JC (Jukes-Cantor), HKY\{2.0\} (the Kimura 2-parameter model with transitions twice as likely as transversions), HKY+GC\{1\} (the Kimura 2-parameter model with variable transition/transversion ratio and substitution rate drawn randomly per site from a continuous Gamma(1,1) distribution), and LG+GC (the Le-Gascuel protein matrix with rates drawn from a continuous Gamma distribution).  For brevity the  HKY+GC\{1\} and LG+GC conditions are referred to in the paper as HKY and LG respectively. Nucleotide frequencies were drawn from common empirical distributions (see \cite{naser-khdour2021} for details).  AA frequencies were drawn from a uniform distribution.

Indels were simulated with insertion and deletion rates equal to $0.01\times$ the substitution rate. Insertion and deletion sizes were drawn from Geometric distributions with mean 5 and 4, respectively.

In all cases training alignments contained 500 characters and test alignments 1000 characters, with the exception of the indel alignments which contained 500 amino acid residues but were of variable length (up to 5000).  Most of the gene trees in RAxMLGrove \cite{hohler_raxml_2022}, the largest public database of inferred gene trees, contain fewer than 50 taxa and are less than 200 characters in length.  This motivated our choice of simulation conditions.

\begin{figure}
    \centering
    \includegraphics[width=0.7\linewidth]{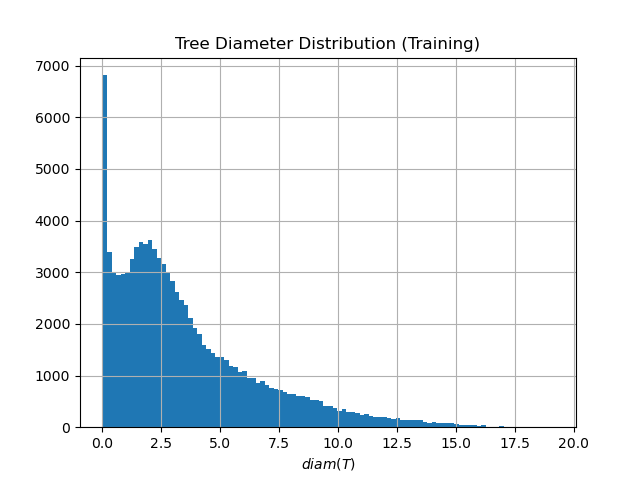}
    \caption{Distribution of tree diameters.}
    \label{fig:diameters}
\end{figure}

The \textit{diameter} of a tree is the maximum patristic distance $d_T$ between any two leaves. Most simulated trees had diameters less than 2.5, with the maximum diameter being $19.145$ (Figure \ref{fig:diameters}).

IQ-TREE trees were inferred using the default parameters and the appropriate model (JC, HKY, HKY+G, or LG+G).  IQ-TREE accepts alignments with gaps (i.e. the indel dataset) but does not have a model for indels: the likelihood of a site having gaps in some sequences is simply calculated as the likelihood of the subtree obtained by excluding those sequences.

\subsection{Model Training}

Models were implemented in PyTorch \cite{paszke2019} and PyTorch Lightning.
All models were trained using the Adam optimizer with $\beta_1=0.9,\beta_2=0.999,\epsilon=10^{-8}$, initial learning rate of 0.01, and cosine learning rate decay.  Models were trained for a maximum of 100 epochs; early stopping (with $d_{RF}$ on a separate 60-taxon validation set as the criterion) was used to prevent overfitting.  Models were trained on a single Tesla V100-PCIE with 32GB of memory.

For each test condition, the DNN's were trained on in-distribution data (e.g. LG+GC for both train and test conditions).

\subsection{Loss Functions on Distance and Covariance Matrices}\label{sec:optim}

In the main text we report results for models trained with the Mean Absolute Error (MAE) loss.  We also explored several other custom loss functions and regularization techniques.

We found that MAE loss was superior to the Mean Squared Error (MSE) loss.  We suspect this is because MAE penalizes outliers to a lesser extent than MSE, and a significant number of training instances lie in the `hard' (possibly unlearnable) regime.  To confirm this hypothesis we retrained using the $L_{2,1}$ matrix norm over batches. This norm penalizes distances on each tree according to MSE, but the weighting \textit{across} trees follows MAE.  We did not find a significant difference between $L_{2,1}$ training and MAE training.

 Training $S_c$ networks presents special challenges.  While the MSE loss can be used on the Gram matrices $C=(Z_i^TZ_j)_{ij}$, it is more appropriate to use losses that are specific to the cone of positive semi-definite matrices which contains $C$.  Such losses include the 
Log-Determinant divergence \cite{Kulis2009,Davis2007}, aka Stein's loss:
\begin{align*}
    D_{ld}(X||Y)\equiv \text{tr}(XY^{-1})-\log\det(XY^{-1})-n
\end{align*}
and the von Neumann divergence
\begin{align*}
    D_{vn}(X||Y)\equiv \text{tr}(X\log X-X\log Y -X+Y)
\end{align*}

These divergences assume $X$ and $Y$ are full rank. This is always the case for the phylogenetic covariance matrix of a binary tree; it is also the case for the learned $C$ matrix provided all input sequences are distinct.

$ D_{ld}$ has some useful properties.  $D_{ld}(X,Y)=D_{ld}(P^{T}XP,P^{T}YP)$ for invertible $P$.  This implies that it is invariant to relabeling of taxa (given by $X\mapsto P^{T}XP$ for some permutation matrix $P$).  Of course, this property is shared by the MSE and MAE losses.  Unlike these losses, LogDet is also invariant to scaling: $D_{ld}(X,Y)=D_{ld}(\alpha X,\alpha Y)$ for any $ \alpha>0$.  For tree metrics, this means it is invariant to ultrametric transformations (scalings which preserve patristic distances). This can be useful when training on heterogeneous datasets (i.e datasets wherein alignments have radically different timescales or mutation rates).
This also makes LogDet an appropriate surrogate loss for the non-differentiable Robinson-Foulds distance, which ignores branch lengths entirely.  

$D_{vn}$ does not require inverting the $Y$ matrix and is therefore much faster to compute and more numerically stable than  $D_{ld}$.

We find that $D_{ld}$ and $D_{vn}$ do learn slightly better $S_c$ networks than MSE.

Von Neumann and LogDet divergence can also be used as regularizers when a distance matrix is the target; this requires transforming the  dissimilarity matrix to a PSD covariance matrix (e.g. via $X\mapsto \exp(-\gamma X)$ for $\gamma>0$).  This can encourage $P$ and $S_d$ networks to learn valid metrics; however as we show these networks often produce metrics without the need for explicit regularization.

Finally, we note that Atteson's theorem depends on the $\ell_\infty$ edge radius of a tree. 
This can be optimized directly using the Boltzmann (smooth max) operator.
We found that employing a smooth-max loss (that is, XXX)  does not lead to significantly better predictors, and in fact can decrease performance by causing the training algorithm to focus too much on ``hard'' trees.  It is still possible that relaxations of this criterion, or other sufficient criteria for NJ reconstruction such as quartet consistency and quartet additivity \cite{mihaescu_why_2009}, may provide useful loss functions for SML.  In particular, contrastive learning \cite{khosla_supervised_2020} could be used to optimize edges for triplets or quartets of taxa.

\subsection{Performance Evaluation}
There are a multitude of distance functions for comparing phylogenetic trees.  In their evaluation of Phyloformer, \cite{nesterenko2025} report the weighted Robinson-Foulds distance, which measures the accuracy of inferred branch lengths in addition to tree topologies.  We found that Attention networks capture branch lengths much better than Invariant networks (and even, in some cases, better than IQ-TREE).  However, this misrepresents the true utility of these models.  Given a topology, accurate branch lengths can be inferred by a variety of methods (e.g. least squares, see below).  Thus, a small gain in branch length accuracy does not justify a large loss in topological accuracy.  The unweighted Robinson-Foulds therefore provides a more honest appraisal of model performance.

\textbf{Robinson-Foulds Distance}.  A binary tree $T$ on leaf set $X$ can be specified by the \textit{splits} it induces on $X$.  Each edge in $T$ partitions X into two disjoint subsets $X_1$ and $X_2$.  Let the set of all such splits be denoted $\Sigma(T)$. The Robinson-Foulds distance between two $X$-trees $T_1$ and $T_2$ is defined as the cardinality of the symmetric difference of their split sets, $d_{RF}(T_1,T_2)\equiv |\Sigma(T_1)\Delta\Sigma(T_2)|$.  In this work we use the normalized distance: \[d_{RF}(T_1,T_2) = \frac{ |\Sigma(T_1)\Delta\Sigma(T_2)|}{ |\Sigma(T_1)\cup\Sigma(T_2)|}\]

 \subsection{Neighbor-Joining}

Because the number of unrooted trees on $n$ leaves grows as $\gO(n^n)$ and the number of trees considered by NJ grows as $\gO(n^3)$, only a minuscule fraction of possible trees are examined for large $n$.  Nevertheless, NJ remains competitive with many parametric inference methods that explore a much larger collection of trees.

BIONJ \cite{Gascuel1997} differs from NJ in that distances are corrected for the variance in each estimate; longer distances are assumed to have higher variance.
We found that NJ and BIONJ produce nearly identical trees for all of the models evaluated in this chapter.

\end{document}